\documentclass[a4paper]{article}

\usepackage{INTERSPEECH2016}

\usepackage{subfig}
\usepackage{graphicx}

\usepackage{siunitx}
\usepackage{dcolumn}

\usepackage{graphicx}
\usepackage{amssymb,amsmath,bm}
\usepackage{textcomp}

\usepackage{mathtools}
\usepackage{amsmath}
\usepackage{amsfonts}

\usepackage{multirow}

\ninept

\title{Environmental Noise Embeddings For Robust Speech Recognition}

\makeatletter
\def\name#1{\gdef\@name{#1\\}}
\makeatother \name{{\em Suyoun Kim$^1$, Bhiksha Raj$^1$, Ian Lane$^1$}}

\address{$^1$Electrical Computer Engineering\\
Carnegie Mellon University\\
  {\small \tt suyoun@cmu.edu, bhiksha@cs.cmu.edu, lane@cmu.edu}
}

\begin{document}

  \maketitle
  \begin{abstract}
We propose a novel deep neural network architecture for speech recognition that explicitly employs knowledge of the background environmental noise within a deep neural network acoustic model. A deep neural network is used to predict the acoustic environment in which the system in being used. The discriminative embedding generated at the bottleneck layer of this network is then concatenated with traditional acoustic features as input to a deep neural network acoustic model. Through a series of experiments on Resource Management, CHiME-3 task, and Aurora4, we show that the proposed approach significantly improves speech recognition accuracy in noisy and highly reverberant environments, outperforming multi-condition training, noise-aware training, i-vector framework, and multi-task learning on both in-domain noise and unseen noise.
  \end{abstract}
  \noindent{\bf Index Terms}: robust speech recognition, noise adaptation

  \section{Introduction}

\begin{figure*}[!t]
\centering
\subfloat[+N\_DNN]{\includegraphics[width=3.2in]{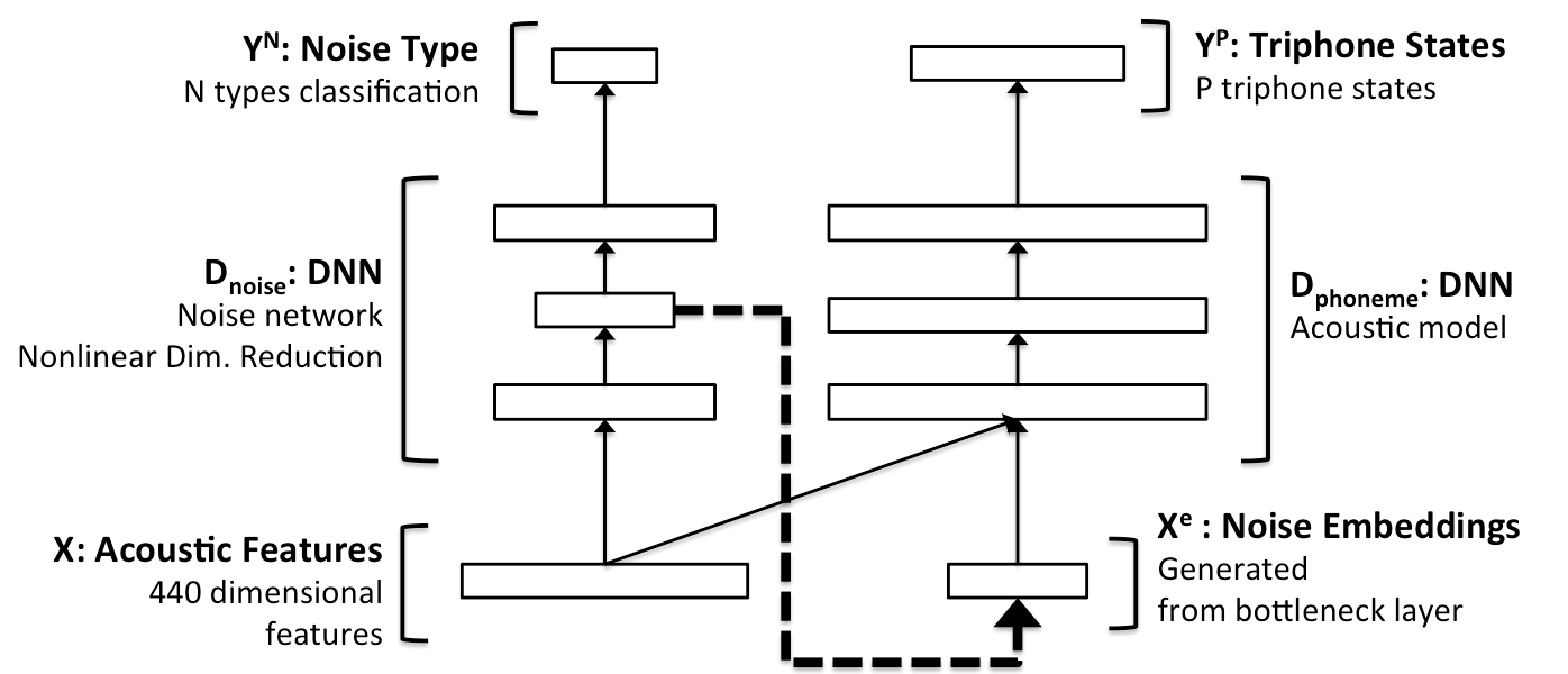} \label{fig:architecture}}
\hfil
\subfloat[MTL]{\includegraphics[width=3.2in]{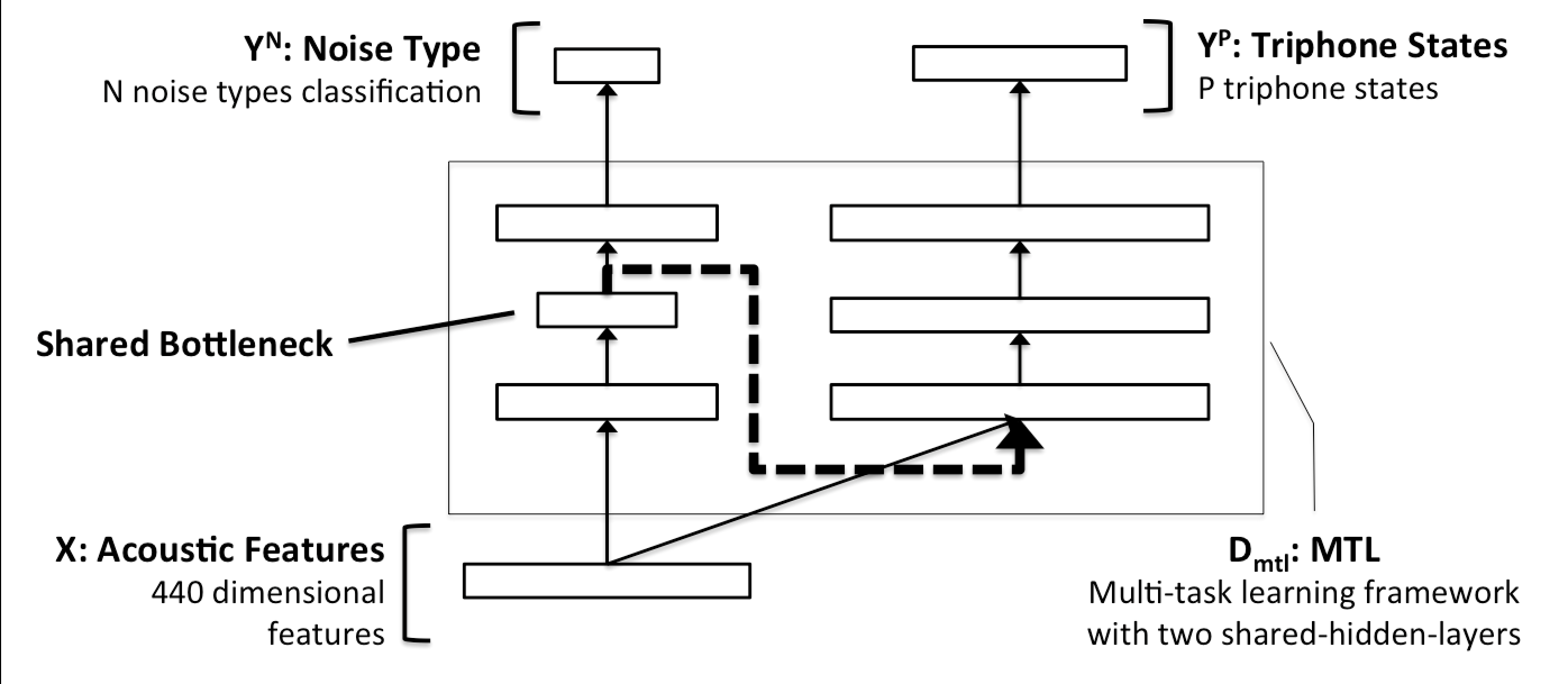} \label{fig:architecture2}}
\caption{Illustration of our approach noise embedding adaptive training \texttt{+N\_DNN} and \texttt{MTL} framework. (a)\texttt{+N\_DNN} is sequentially training two parts of the same network: (1) train environmental embeddings, then (2) train the triphone network. By contrast, (b)\texttt{MTL} is jointly optimized the two components of the network.} 
\end{figure*}  

In many speech recognition tasks, despite an increase in the variability of the training data, it is still common to have significant mismatches between test environment and training environment, e.g. ambient noise and reverberation. This environmental distortion results in the performance degradation of automatic speech recognition (ASR). Various techniques have been introduced for increasing robustness in this situation.

Over the years, prior works on improving robustness under environmental distortion has generally fallen into three categories: feature enhancement, transformation, and augmentation with auxiliary information. Feature enhancement approaches try to attenuate the corrupting noise in the observation and develop more robust feature representation in order to minimize the mismatches between training and test conditions. Many of these methods have been proposed to suppress noise, for example, the model-based compensation methods, Vector Taylor Series (VTS), attempt to model the nonlinear environment function and then apply the compensation for the effects of noise \cite{moreno1996vector}, the noise robust feature extraction algorithms based on the different characteristics of speech and background noise have been developed \cite{kim2012power, kim2010nonlinear}, and the missing feature approaches, attempt to mask or impute the unreliable regions of the spectral components because of degradation due to noise have been proposed \cite{raj2005missing, li2013improving, narayanan2014joint}. Transformation approaches attempt to transform the feature or model space adaptively according to each speaker or each utterance \cite{gales1998maximum, gales1999semi}. 

One recent approach involves augmenting the acoustic features with auxiliary information that characterizes the testing conditions, such as a noise estimates \cite{seltzer2013investigation}. This approach attempts to enable the Deep Neural Network acoustic model \cite{mohamed2012acoustic,hinton2012deep, seide2011conversational} to learn the relationship between noisy speech and noise directly from the data by giving additional cues. Instead of providing the preprocessed or normalized feature to the network, the network figures out the normalization during training by using its exceptional modeling power. In order to do that, the network is informed by the noise identity features. The Noise-Aware Training (NAT) proposed in \cite{seltzer2013investigation} uses an estimate of the noise for the noise identity feature. In this work we extend the prior work, NAT, with an improved method to model and represent dynamic environmental noise.

Related work includes the use of identity vector (i-vector) representation based on the Gaussian Mixture Models (GMMs). The i-vector is a popular technique for speaker verification and speaker recognition, and it captures the acoustic characteristics of a speaker's identity in a low-dimensional fixed-length representation. For this reason, it has been used for speaker adaptation in ASR \cite{liu2015investigation, saon2013speaker}. However, the i-vector framework has only been applied to speaker adaptation, not to noise adaptation. The success of the i-vector framework in speaker adaptation of DNN acoustic models motivated us to look at their applicability to noise adaptation. 

In this work, we propose a noise adaptation framework that can dynamically adapt to various testing environments. Our framework incorporates environmental acoustics during the DNN acoustic model to improve robustness in environmental distortion. The model explicitly employs knowledge of the background noise and learns the low-dimensional noise feature from the discriminatively trained DNN, which we call \textit{noise embeddings}. Through a series of experiments on Resource Management (RM) \cite{rmdata}, CHiME-3 task \cite{chime3}, and Aurora4 datasets \cite{parihar2002aurora}, we show that our proposed approach improves speech recognition accuracy in various types of noisy environments. In addition, we also compare our approach with the NAT \cite{seltzer2013investigation}, the i-vector framework \cite{madikeri2015integrating,saon2013speaker}, and a multi-task learning framework that jointly predicts noise type and context-dependent triphone states.

The paper is organized as follows. In Section \ref{sec:method} we review other noise adaptation systems, NAT, i-vector framework and our proposed noise adaptation framework. In Section \ref{sec:exp}, we evaluate the performance of the proposed approach. Finally, we draw conclusions and discuss future work in Section \ref{sec:conclusion}.

\section{Environmental Noise Adaptation}
\label{sec:method}

\subsection{Noise Aware Training}
One framework that has been used for the noise adaptation is Noise-Aware Training (NAT) which is proposed in \cite{seltzer2013investigation}. NAT is designed to make the DNN acoustic model automatically learn the relationship between each observed input and the noise present in the signal by augmenting an additional cue, the noise estimates. This noise estimate is simply computed by averaging the first and last ten frames of each utterance. The NAT achieves approximately 2\% relative improvement in word error rate (WER) evaluating on the Aurora4 dataset \cite{parihar2002aurora}. However, as the NAT assumes the noise is stationary and uses a noise estimate that is fixed over the utterance, the performance of this technique relies on the characteristic of the background noise and prior knowledge of the region of the noisy frame. In this work, we explore a way to represent the noise to improve adaptation performance.

\subsection{Identity Vector for Noise}
The i-vector framework is a popular technique for speaker recognition and it captures the acoustic characteristics of a speaker's identity in a low-dimensional fixed-length representation. From this reason, it has been used as a speaker adaptation technique for ASR and consistently achieves 5-6\% relevant improvement in WER(\%). The success of the i-vector framework in speaker adaptation of DNN acoustic models motivated us to look at their applicability to noise adaptation. 

Here we review the main idea behind the i-vector framework. The acoustic feature vectors $\mathbf x_t \in {\rm I\!R^D}$ are seen as samples generated from a universal background model (UBM) represented as a GMM with $K$ diagonal covariance Gaussians. The key of the i-vector algorithm is to assume a linear dependence between the speaker-adapted with respect to the UBM, supervector $\mathbf s$, and the speaker-independent, the mean of supervectors, $\mathbf m$:

\begin{equation}
\mathbf{s} = \mathbf{m} + \mathbf{Tw}
\end{equation}

where $\mathbf T$ of size $D$ x $M$, is the factor loading submatrix corresponding to component $k$ and $\mathbf w$ is the size of the $M$ speaker identity vector (i-vector) corresponding to speaker. We estimate the posterior distribution of $\mathbf w$ given speaker $s$ data $\mathbf x_t(s)$ using the EM algorithm. The i-vector extraction transforms are estimated iteratively by alternating between evaluating $\mathbf w$ in E step and updating the model parameters $\mathbf{T}$ in M step.

In this work, instead of using the speaker ID in the general application of the i-vector system, we used the noise type for generating noise i-vector. 

\subsection{Learning environmental noise embeddings}
\label{sec:ndnn}

\begin{figure*}[t]
  \centering
  \includegraphics[width=.23\linewidth]{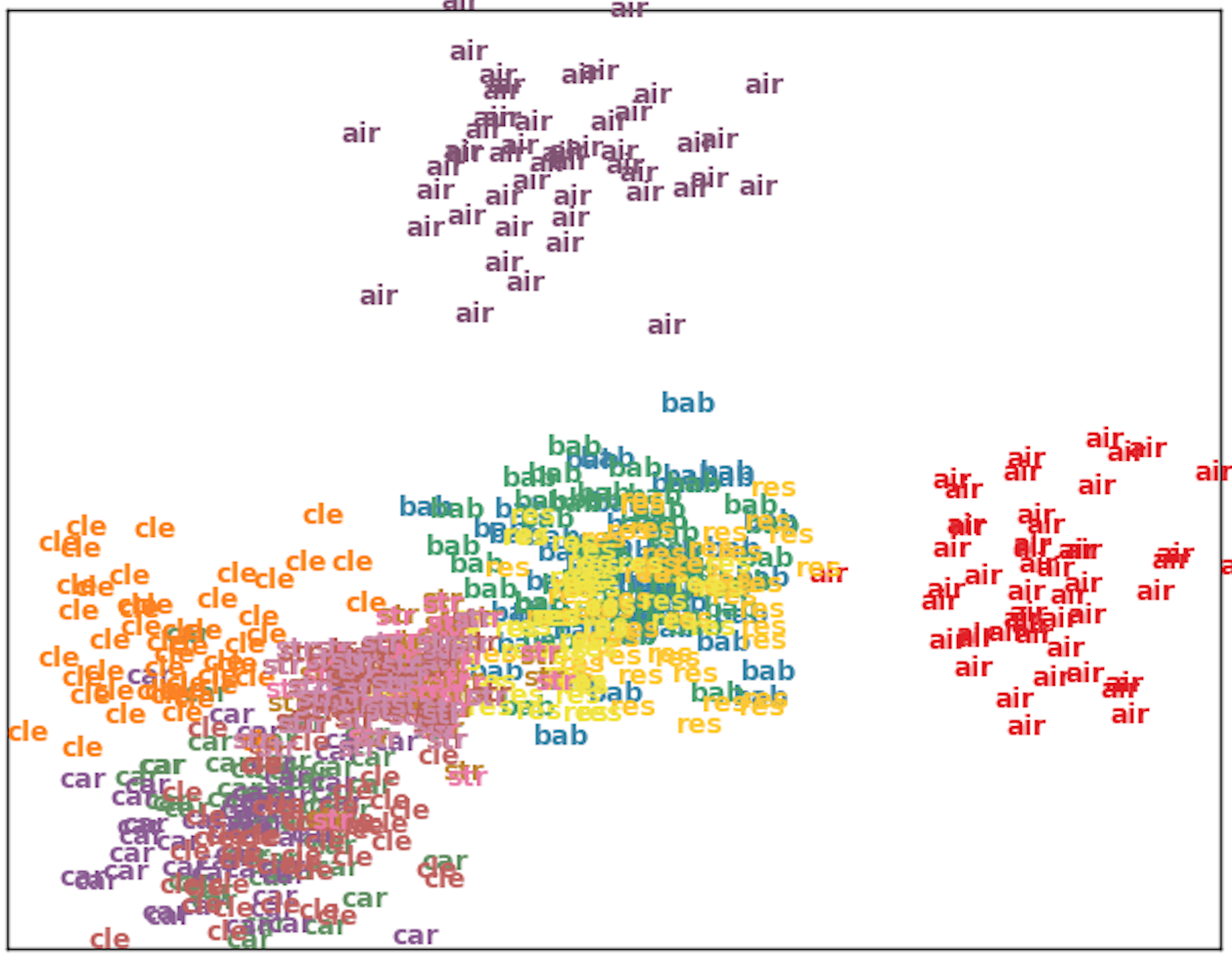}
  \includegraphics[width=.23\linewidth]{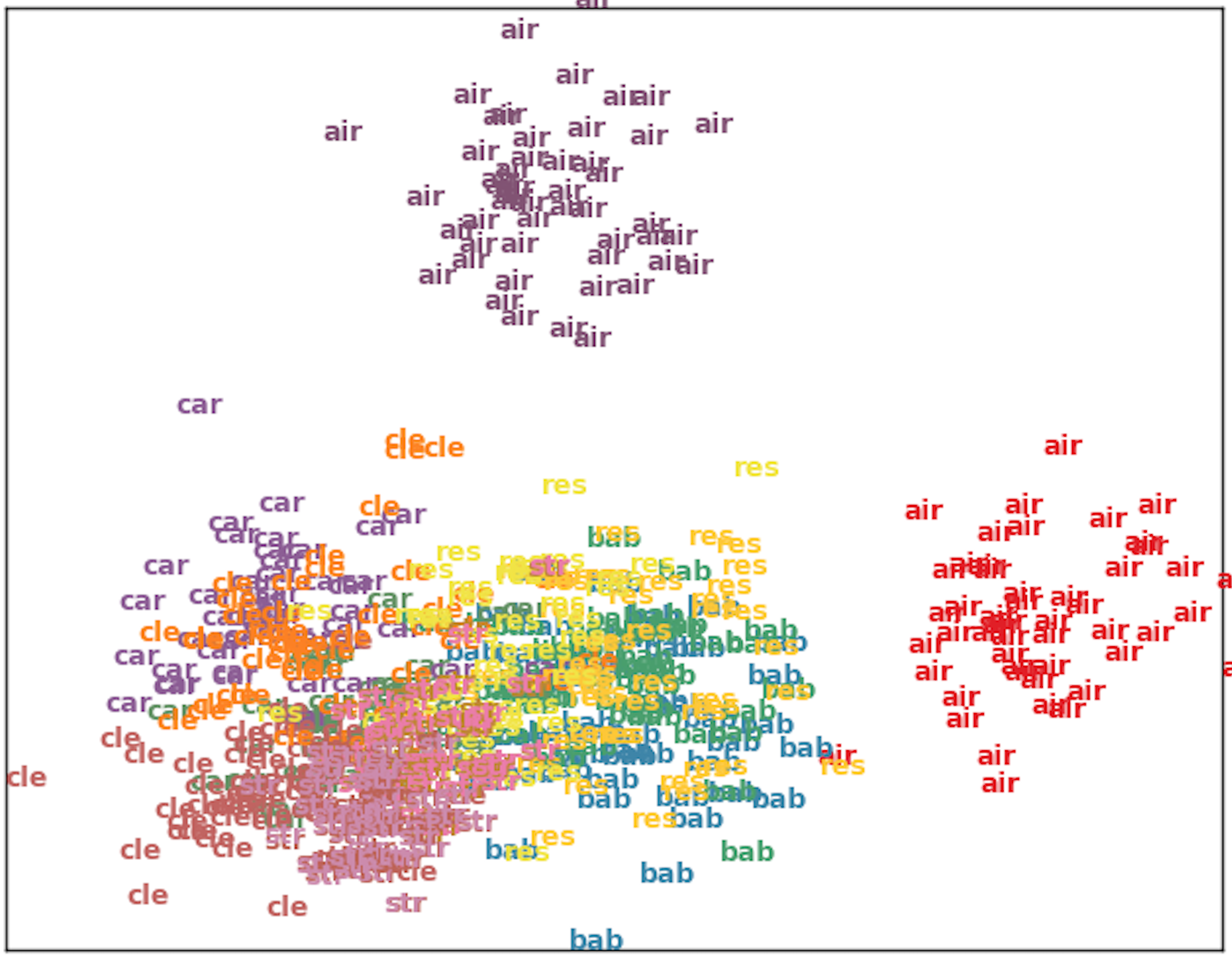}
  \includegraphics[width=.23\linewidth]{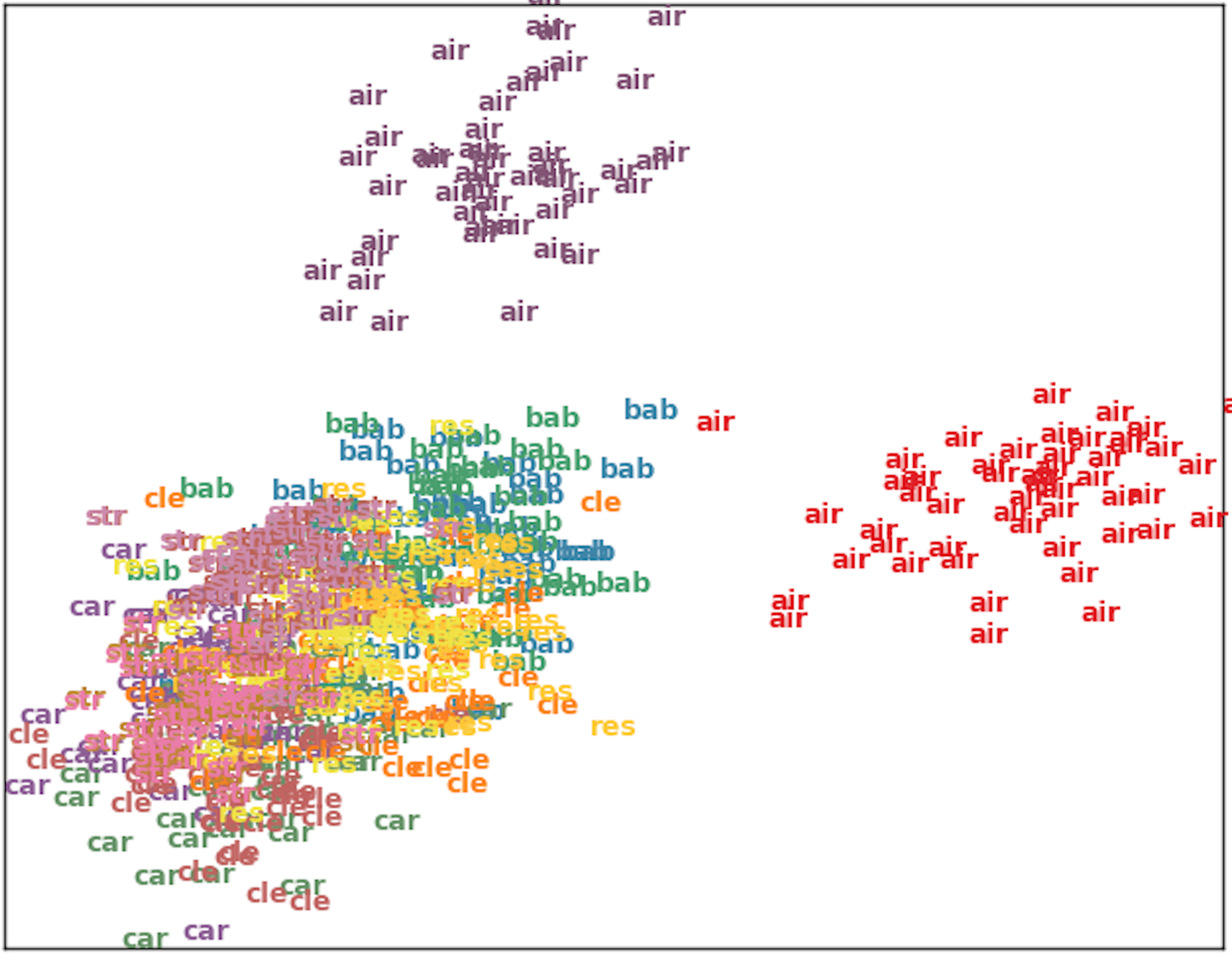}
  \includegraphics[width=.23\linewidth]{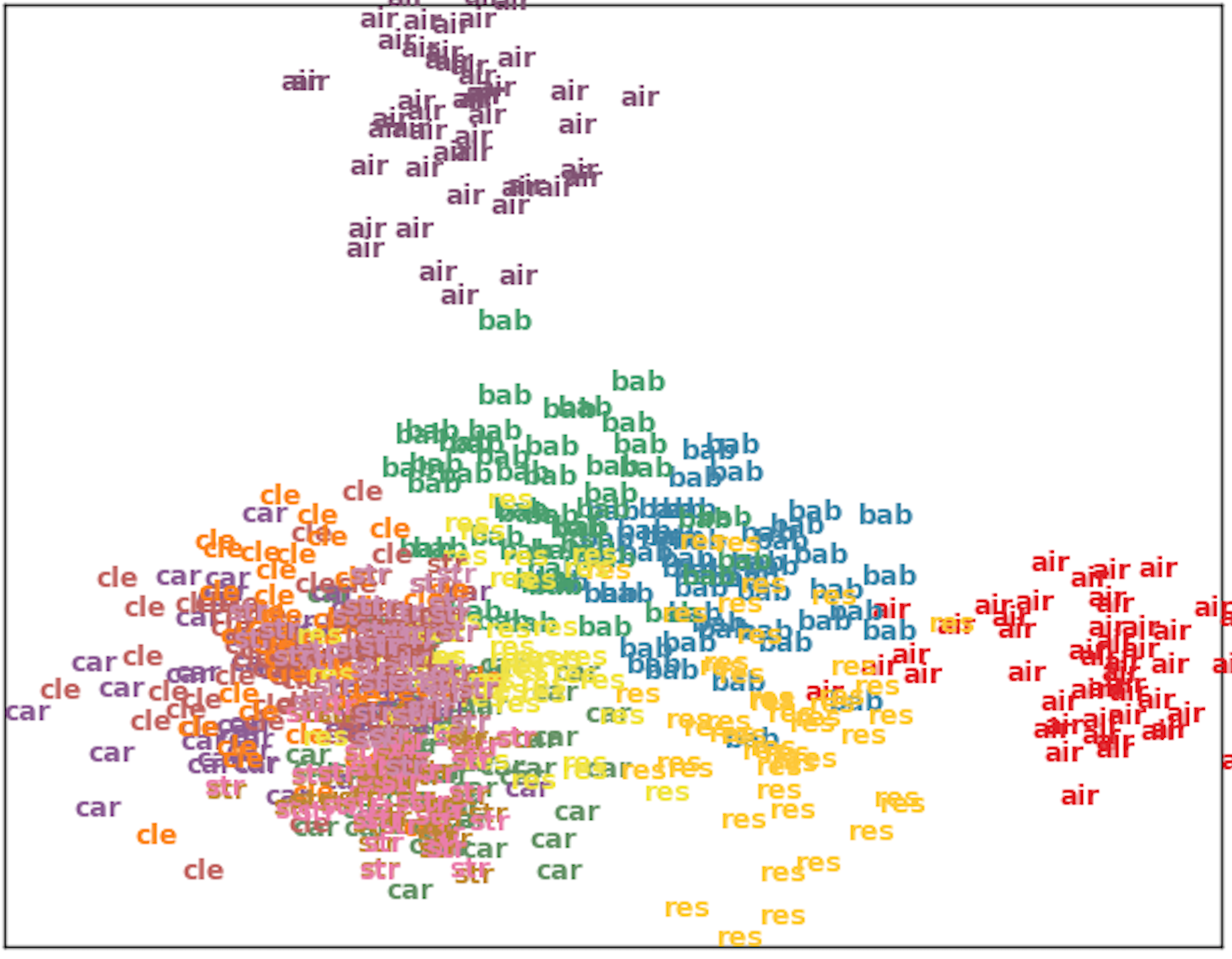}
\caption{A comparison of the final input features of the unseen noise set, Aurora4 evaluation \cite{parihar2002aurora}, from the different algorithms \texttt{baseline}, \texttt{+N\_NAT}, \texttt{+N\_GMM}, and \texttt{+N\_DNN}. The randomly selected 700 input features projected in 2-dimensional space by LDA. The 40-dimensional noise features generated from the model trained on CHiME-3 training set were augmented. The colors represent each type of noise condition.
}
\label{fig:lda}
\end{figure*} 

In this subsection we describe our approach, which explicitly employs knowledge of the background environmental noise within a DNN acoustic model to improve robustness under environmental distortion. Our approach is motivated by previous work on NAT, and extends the way of representing the noise adaptation data. Unlike NAT, our system can dynamically adapt to different testing environments by appending varying noise estimates at each input frame. 

Our proposed system consists of two subnetworks with different objectives for each. As shown in Figure \ref{fig:architecture}, the left $D_{noise}$ learns the noise embeddings and the right $D_{phoneme}$ is the regular acoustic model. The networks are optimized sequentially.

First, we learn the noise embeddings at each frame from a narrow bottleneck hidden layer in $D_{noise}$, given various types of noisy speech data. We start with training $D_{noise}$ with the regular acoustic feature, $X$, to classify the different ground-truth categorical labels, the noise types, $Y^N$. We use a bottleneck neural network for $D_{noise}$. A bottleneck neural network is a kind of multi-layer perceptron (MLP) in which one of the internal layers has a small number of hidden units, relative to the size of the other layers. The common approach to extracting the feature vectors is to use the activations of the bottleneck hidden units as features \cite{grezl2007probabilistic}. It has been shown that the features generated from the bottleneck network can be classified into a low-dimensional representation by forcing this small layer to create a constriction in the network. Consequently it can be represented as a nonlinear transformation and leads to dimensionality reduction of the input features. We take advantage of this fact to generate the low-dimensional secondary feature vector. To make the bottleneck feature vector embed the discriminative acoustic characteristics of background noise instead of the phonetic characteristics, the task of the network is to classify different noise conditions. 

Once the $D_{noise}$ is optimized, we extract the noise embeddings $X^e$ at each input frame from the bottleneck hidden layer in $D_{noise}$. The learned noise embeddings $X^e$ are then concatenated to each corresponding original acoustic feature frame. The noise estimates keep changing over the time frame; our noise adaptation technique does not require the assumption that the noise is stationary.

Finally, we train $D_{phoneme}$ with input features $X$ and $X^e$ to classify the phonetic states, $Y^P$, as in usual acoustic modeling. In the decoding step, the noise label is not required and we can obtain the noise embedding by forwarding the acoustic features to the optimized $D_{noise}$. The Figure \ref{fig:architecture} illustrates the overall architecture. 

\subsection{Multi-task learning}
We recognize that our framework described in Section \ref{sec:ndnn} is sequentially training two parts of the same network. First we train the environmental embeddings, and then we fix it and train the triphone network. As a comparator, we also attempt joint optimization. Here the two components of the network are jointly optimized. This joint optimization approach can be effectively a multi-task learning setup which is a method that jointly learns more than one problem together at the same time using shared representation. It has been applied to various speech-related tasks, and our setup \texttt{MTL} is similar to these other multi-task learning solutions \cite{seltzer2013multi}, except that we are considering environment as the variable. 

Figure \ref{fig:architecture2} shows the architecture of our \texttt{MTL} approach. We jointly optimize the network to predict the noise label while to predict the triphone states, so that the network can learn noise-related structure. As a secondary task, the noise label classification task is designed to predict the acoustic environmental type $Y^{N}$ from the current acoustic observation $\mathbf X$. For the fair comparison to our framework, \texttt{+N\_DNN}, we build the same size of the network in which the two hidden layers are shared across two different task. Especially we make the second shared-hidden-layer has the same dimension as that of our noise embedding feature, so that this second shared-hidden-layer can serve as environmental noise information. Once the network is optimized to minimize both the noise prediction error and the triphone states error, two shared-hidden-layers and the right side of three hidden layers are used for the decoding. 


\section{Experiments}
\label{sec:exp}
\subsection{Dataset}
We investigate the performance of our noise embedding technique on three different databases, RM \cite{rmdata}, CHiME-3 task \cite{chime3}, Aurora4 \cite{parihar2002aurora}, in two main ways: in-domain noise experiment, and unseen experiment. In-domain noise experiment, we perform the experiments on the test set with the same types of noises when the model is trained. For the unseen noise test, we trained the model on the CHiME-3 dataset, and then tested it with the evaluation set of the Aurora4 task. 

We first evaluated our method on the in-domain experiments on the noisy data that have been derived from RM. We artificially mixed the clean speech with eight different types of noisy background, including: white noise at 0 dB, and 10 dB SNR, street noise at 0 dB, and 10 dB SNR, background music at 0 dB, and 10 dB, and simulated reverberation with 1.0 s reverberation time and 600 ms reverberation time. The street noise and the background music segments was obtained from \cite{kim2012power}, and the reverberation simulations were accomplished using the \textit{Room Impulse Response} open source package \cite{habets2006room}, and the virtual room size was 5 x 4 x 6 meters.

The CHiME-3 challenge task includes speech data that is recorded in real noisy environments (on a bus, in a cafe, in a pedestrian area, and at a street junction). The training set has 8,738 noisy utterances (18 hours), the development set has 3,280 noisy utterances (5.6 hours), and the test set has 2,640 noisy utterances (4.5 hours). 

The evaluation set of Aurora4 task consists of 9.4 hours of 4,620 noisy utterances corrupted by one of 14 different noise types, which combine 7 different background noise types (street traffic, train station, car, babble, restaurant, airport, and clean) and 2 channel distortions. The noise adaptation features for the Aurora4 task were extracted from the network optimized on the CHiME3 training set without any of the environment information of the Aurora4 task.

We followed the standard way of representing speech by using Kaldi toolkit \cite{Povey_ASRU2011} with their standard recipe. Every +5 and -5 consecutive MFCC feature frames are spliced together and projected down to 40 dimensions using LDA, then fMLLR transform is computed on top of the features.

\subsection{System training}
To evaluate the proposed techniques, we built six different systems: \texttt{baseline}, noise-aware-training \texttt{+N\_NAT}, the offline i-vector framework \texttt{+N\_GMM}, the online i-vector framework \texttt{+N\_GMM\_ON}, our proposed system, \texttt{+N\_DNN} and \texttt{MTL}.

For our \texttt{baseline}, we trained the DNN acoustic model without any auxiliary adaptation data. The network contains 7 hidden layers that have 2,048 units each. We trained the network using the cross-entropy objective with mini-batch based stochastic gradient descent (SGD). We followed the same baseline pipeline provided by the CHiME-3 organizer \cite{chime3} and matched up WER with the official baseline.  

For \texttt{+N\_NAT}, we estimated the noise the same way as previous work \cite{seltzer2013investigation}. We simply averaged the first and last ten frames of each utterance, creating an estimate that was fixed over the utterance. 

For another comparator \texttt{+N\_GMM} and \texttt{+N\_GMM\_ON}, we followed the standard offline and the online i-vector extraction method \cite{saon2013speaker,madikeri2015integrating}. We built a Universal Background Model (UBM) using 2,048 Gaussians and extracted a 40 dimensional i-vector of the corresponding noise type. For online i-vector, we use 10 frames of speech as a window.

For our proposed model \texttt{+N\_DNN}, we built a DNN that has a narrow bottleneck hidden layer, allowing for the extraction of more tractable, high-level noise context information. It has five hidden layers. The fourth layer is a bottleneck with 40 units. Other layers have 1024 units each. Once the network was optimized, the discriminative noise features of every training and test set were concatenated to each corresponding original feature set. Unlike previous noise estimates \cite{seltzer2013investigation}, our noise features were focused on capturing the background information optimized by different objectives, classifying the noise types, and estimating every input frames without assuming that the noise is stationary. 

For the multi-task learning system, \texttt{MTL}, we shared two layers as described in Figure \ref{fig:architecture2}. For the fair comparison, the number of model parameters are matched approximately.

\subsection{Results}
\label{sec:results}

\begin{figure}[htb]
  \centering
  \includegraphics[width=0.8\linewidth]{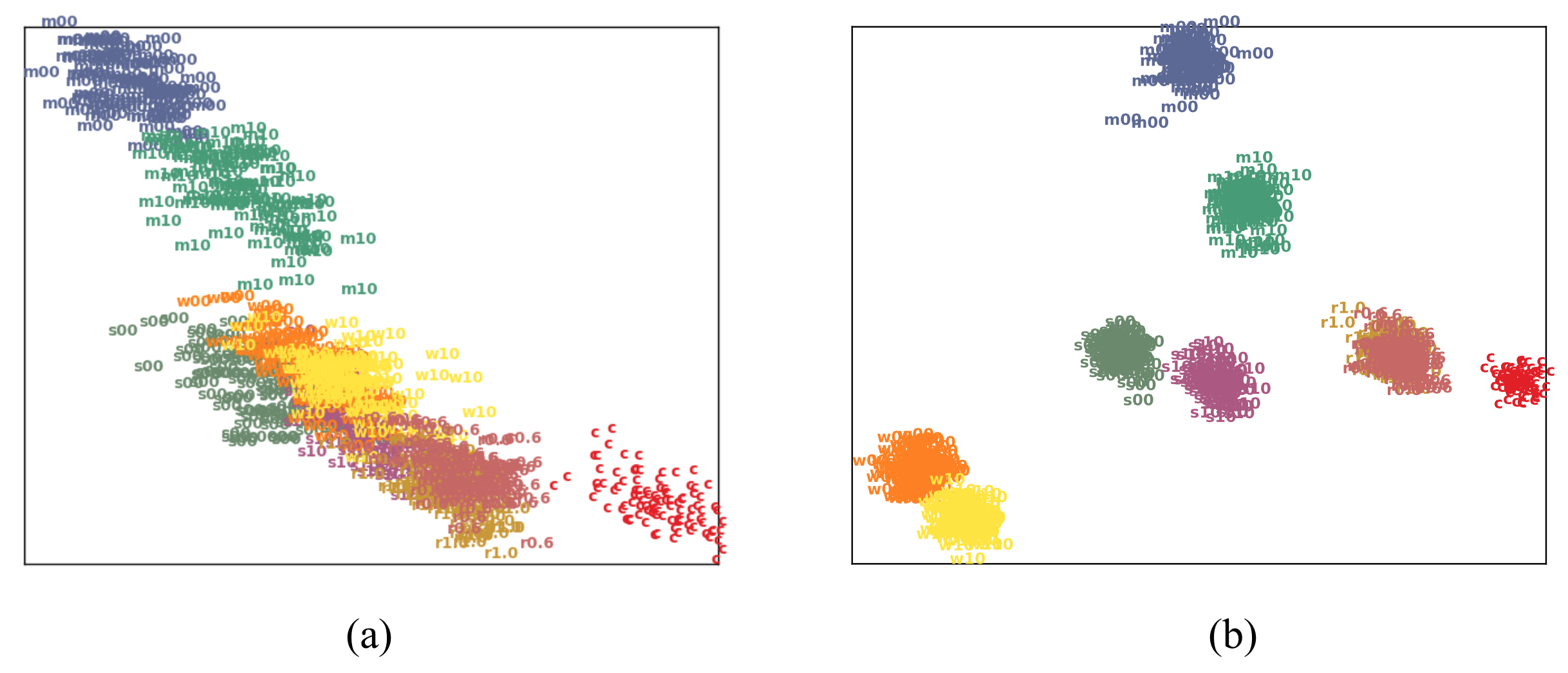}
  \caption{Comparison of the final input features of in-domain noise (RM) between \texttt{baseline} and \texttt{+N\_DNN}. The randomly selected 100 input features projected in 2-dimensional space by LDA.}
  \label{fig:embedding}
\end{figure} 

Before we evaluated the recognition accuracy, we first visualized the final input features of different systems. Figure \ref{fig:embedding} shows the final input feature of in-domain noise set (RM) of \texttt{baseline} and \texttt{N\_DNN}. The figure shows that adding noise embeddings helps the input feature set be significantly more discriminative with respect to the different environments. Figure \ref{fig:lda} shows the final input feature of unseen noise set (Aurora4 evaluation set) of \texttt{baseline}, \texttt{+N\_NAT}, \texttt{+N\_GMM}, and \texttt{+N\_DNN}. The figure shows that the input features augmented with the noise feature based on \texttt{+N\_DNN} are relatively more discriminative with respect to the different environments and it indicates that the model is work well on even unseen noise case. 

Table \ref{tab:rmresult} compares the recognition accuracy obtained using three models: \texttt{baseline}, \texttt{MTL}, and \texttt{+N\_DNN}. It can be seen that at all SNRs and all noise types \texttt{+N\_DNN} outperforms the others even in clean datasets. We note that the improvements in recognition accuracy are greater at the lower SNRs. For example, we obtained 2.92 \% of WER improvement in the dataset with background music at 0 dB SNR, whereas only 0.19 \% of WER improvement in the clean dataset.

\sisetup{table-format = 2.1}
\begin{table}[!t]
\caption{Comparison of WERs(\%) between the \texttt{baseline}, \texttt{N\_DNN}, and \texttt{MTL} model using 50-dimensional embeddings for 8 different noisy evaluation sets and one clean evaluation set.}
\label{tab:rmresult}
\centering
\resizebox{0.65\columnwidth}{!}{
\begin{tabular}{c||c|c|c}
\hline
 \textbf{Testset(SNR/RT)} &  \textbf{\texttt{baseline}} &   \textbf{\texttt{+N\_DNN}} &  \textbf{\texttt{MLT}}\\
\hline\hline
clean       & {3.0}    &  \textbf{2.9}   & 3.1   \\ \hline
music(00)   & {28.4}   &  \textbf{25.5}  & 29.1  \\ \hline
music(10)   & {6.5}  &  \textbf{6.3}   & 7.4    \\ \hline
reverb(0.6) & {16.4}   &  \textbf{15.4}  & 17.4   \\ \hline
reverb(1.0) & {26.8}   &  \textbf{25.3}  & 29.0  \\ \hline
street(00)  & {35.0}   &  \textbf{32.7}  & 39.1  \\ \hline
street(10)  & {7.7}    &  \textbf{6.7}   & 7.7    \\ \hline
white(00)   & {30.7}   &  \textbf{28.8}  & 33.8  \\ \hline
white(10)   & {9.7}    &  \textbf{8.3}   & 9.5  \\ \hline
Average     & {18.3}   &  \textbf{16.9}   & 19.5  \\ \hline
\end{tabular}
}
\end{table}

Table \ref{tab:result_chime3} compares the WER obtained using \texttt{Baseline}, \texttt{+N\_GMM}, \texttt{+N\_NAT}, and \texttt{+N\_DNN}. We note that our approach \texttt{+N\_DNN} provided an additional 2.2\% relative reduction in WER compared to \texttt{Baseline}. Also, it can be seen that the performance of \texttt{+N\_NAT} is highly relies on the dataset and it does not work on CHiME-3 task. Unlike speaker adaptation results, the \texttt{+N\_GMM} showed worse performance than even \texttt{Baseline}. This result is due to insufficient noise diversity in noise i-vector training whereas relatively more available speaker diversity (e.g. 87 speakers are available in CHiME-3 task)

The right-most column in Table \ref{tab:result_chime3} shows WER obtained using \texttt{Baseline}, \texttt{+N\_NAT}, \texttt{+N\_GMM}, \texttt{+N\_GMM\_ON}, and \texttt{+N\_DNN}. Although the improvement of the unseen noise case (relative improvement: 0.9\%) is less than the gain of the in-domain noise case (relative improvement: 2.2\%), it is clear that our noise adaptation approach \texttt{+N\_DNN} is superior to other noise adaptation techniques. This result is also due to insufficient noise diversity, so we expect further improvement can be achieved by using additional noise types during model training. Also, \texttt{+N\_NAT} (12.6\%) and \texttt{+N\_GMM} (12.4\%) are worse than \texttt{Baseline} and this result suggests that our proposed system could be more robust adaptation technique even when the test environments are mostly unknown.

\begin{table}[!t]
\caption{Comparison of WERs(\%) on the CHiME-3 task (In-domain Noise 4.5hrs) and the Aurora4 task (Unseen Noise 9.4hrs) between the \texttt{baseline}, \texttt{+N\_NAT}, \texttt{+N\_GMM}, \texttt{+N\_GMM\_ON}, and \texttt{+N\_DNN}. 40 dimensional noise embeddingss were augmented for noise adaptation. The models are trained on CHiME-3 training dataset (18hrs). (*) denotes the statistical significance ($\alpha = 0.05$) \cite{gillick1989some}.}
\label{tab:result_chime3}
\resizebox{\columnwidth}{!}{
\begin{tabular}{c||c |c||c}
  \hline
\multirow{2}{*}{\bf{Model (CHiME-3)}} & \multicolumn{2}{c||}{\bf{In-domain Noise (CHiME-3)}} & \bf{Unseen Noise (Aurora4)} \\ \cline{2-4} 
                                 & \hspace{5mm}Dev (\%)\hspace{5mm} & Eval (\%) & test\_eval92 (\%) \\ \hline
\texttt{Baseline} & 8.9 & 15.6 & 11.7\\ \hline
\texttt{+N\_NAT }  & \textbf{8.8} & 15.9* & 12.6*  \\ \hline
\texttt{+N\_GMM }  & 8.8 & 15.7 & 12.4*  \\ \hline
\texttt{+N\_GMM\_ON }  & 8.9 & 15.7 & 11.6*  \\ \hline
\texttt{+N\_DNN }  & 8.8 & \textbf{15.3}* & \textbf{11.5}* \\ \hline
  \end{tabular}
}
\end{table}

\section{Conclusions}
\label{sec:conclusion}
We proposed a novel noise adaptation approach, \texttt{N\_DNN}, in which we train a Deep Neural Network that dynamically adapts the speech recognition system to the environment in which it is being used. We verified the effectiveness of our proposed framework with improved recognition accuracy in noisy environments. We also compared our approach to offline and online i-vector framework \texttt{N\_GMM}, \texttt{N\_GMM\_ON}, the Noise-Aware Training, \texttt{N\_NAT}, and \texttt{MTL}. Through a series of experiments on CHiME-3 task and Aurora4 task, we showed our model consistently improves the performance on both in-domain and unseen noise tests with using only four different noise types during training.

In future work, we would scale learning across various noisy data types. We believe further performance improvement even in unseen noisy environments can be achieved by using additional and more diverse noises to cover a wider range of noise variation.

\section{Acknowledgment}
The authors would like to acknowledge the contributions made by Richard M. Stern for his valuable and constructive suggestions during the planning and development of this project.

  \newpage
  \eightpt
  \bibliographystyle{IEEEtran}

  \bibliography{mybib}


\end{document}